\documentclass[letterpaper, 10 pt, conference]{ieeeconf}
\IEEEoverridecommandlockouts
\usepackage{amsmath,amssymb,amsfonts}
\usepackage{algorithmic}
\usepackage{graphicx}
\usepackage{textcomp}
\usepackage{xcolor}
\usepackage{cite}
\usepackage{graphicx}
\usepackage{caption}
\usepackage{multirow}
\usepackage{tabularx}
\usepackage{svg}
\usepackage{pdfpages}
\usepackage{pgfplots}
\usepackage[hyphens,spaces,obeyspaces]{url}
\pgfplotsset{compat=1.8}
\usepgfplotslibrary{statistics}
\usepackage{float}
\usepackage{subcaption}
\usepackage{balance}

\title{\LARGE \bf Robotic Grasping of Harvested Tomato Trusses Using Vision and Online Learning}

\author{Luuk van den Bent,$^{1}$ Tom\'as Coleman$^{1}$ and Robert Babu\v{s}ka$^{2}$
\thanks{$^{1}$Luuk van den Bent and Tom\'as Coleman are with the Department of Cognitive Robotics,
        Delft University of Technology, 2628 CD Delft, The Netherlands,
        {luukbent@gmail.com, t.coleman@tudelft.nl.}}%
\thanks{$^{2}$Robert Babu\v{s}ka is with the Department of Cognitive Robotics, Delft University of Technology, 2628 CD Delft, The Netherlands and with the Czech Institute of Informatics, Robotics, and Cybernetics, Czech Technical University in
Prague, Czech Republic,
{r.babuska@tudelft.nl.}}%
}

\usepackage{tikz}
\usepackage{textcomp}
\usepackage{hyperref}
\usepackage{lipsum}

\newcommand\copyrighttext{%
  \footnotesize \textcopyright 2024 IEEE. Personal use of this material is permitted.
  Permission from IEEE must be obtained for all other uses, in any current or future 
  media, including reprinting/republishing this material for advertising or promotional 
  purposes, creating new collective works, for resale or redistribution to servers or 
  lists, or reuse of any copyrighted component of this work in other works. 
  DOI: \href{https://doi.org/10.1109/ICRA57147.2024.10610089}{10.1109/ICRA57147.2024.10610089}}
\newcommand\copyrightnotice{%
\begin{tikzpicture}[remember picture,overlay]
\node[anchor=south,yshift=10pt] at (current page.south) {\fbox{\parbox{\dimexpr\textwidth-\fboxsep-\fboxrule\relax}{\copyrighttext}}};
\end{tikzpicture}%
}

\begin{document}
%

\maketitle
\copyrightnotice

\begin{abstract}
Currently, truss tomato weighing and packaging require significant manual work. The main obstacle to automation lies in the difficulty of developing a reliable robotic grasping system for already harvested trusses. We propose a method to grasp trusses that are stacked in a crate with considerable clutter, which is how they are commonly stored and transported after harvest. The method consists of a deep learning-based vision system to first identify the individual trusses in the crate and then determine a suitable grasping location on the stem. To this end, we have introduced a grasp pose ranking algorithm with online learning capabilities. After selecting the most promising grasp pose, the robot executes a pinch grasp without needing touch sensors or geometric models. Lab experiments with a robotic manipulator equipped with an eye-in-hand RGB-D camera showed a 100\% clearance rate when tasked to pick all trusses from a pile. 93\% of the trusses were successfully grasped on the first try, while the remaining 7\% required more attempts.
\end{abstract}

\section{INTRODUCTION}
\label{sec:introduction}

During the last decades, crop production has significantly increased in volume and efficiency thanks to mechanization and automation \cite{automation}. However, a substantial amount of manual work is still required in the difficult-to-automate processes such as crop harvesting, manipulation or packaging. This presents a serious problem, given the rising demand for food and the decreasing number of people willing to work in agriculture \cite{labour_shortage}. This paper focuses on the automated handling of truss tomatoes, also known as vine tomatoes. A tomato truss refers to the bundle of tomatoes that are still attached to the fruiting stem after harvesting.

\begin{figure}[htpb]
    \centering
    \includegraphics[angle=0, width=0.45\columnwidth]
    {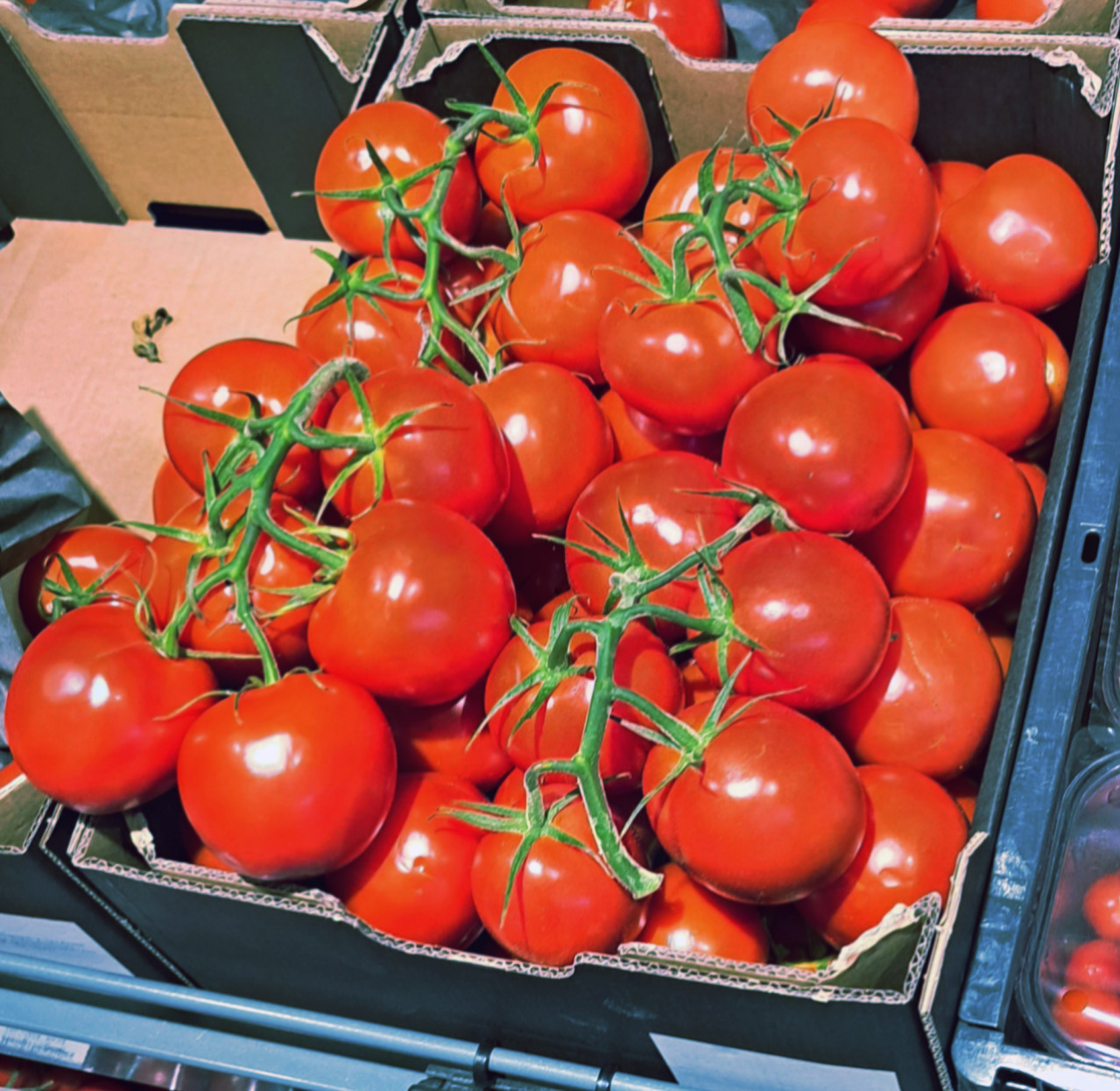}
    \caption{\small Harvested tomato trusses are stacked in a crate before they enter the packaging process.}
    \label{fig:tomato_truss_crate}
\vspace*{-.5cm}
\end{figure}

We focus on grasping trusses from a crate where they are transported from the harvesting location; see Fig.~\ref{fig:tomato_truss_crate}. The purpose is to inspect the tomatoes for damage, weigh them, and finally place them on a transportation belt for automatic packaging. The main challenge is to identify a suitable grasping pose, given the trusses' diverse and unpredictable shapes and the cluttered conditions in the crate. The grasping pose must guarantee safe handling of the tomato truss without damaging it.

This work's main contribution is the development, implementation and validation of a perception method to identify suitable grasp poses so that the trusses can be reliably grasped. We introduce a learning-based grasp pose ranking algorithm to select the most suitable grasp pose out of several candidate poses and to adapt the selection model based on the success or failure of the executed grasp. Extensive lab experiments have been carried out to validate the approach using a Franka Emika Panda manipulator equipped with the Intel Realsense D405 RGB-D (red-green-blue-depth) camera. More than 1300 grasp attempts have been carried out within these experiments on real tomato trusses. The data acquired have been used to develop and train the deep-learning models and to validate the approach. To the best of our knowledge, such extensive truss tomato grasping experiments have never been documented in the literature.

The remaining sections of this paper are structured as follows: Section~\ref{sec:related_work} provides an overview of the related research on grasping tomato trusses. Section~\ref{sec:perception} describes the proposed perception method for finding suitable grasp poses. Section~\ref{sec:experiments} reports validation experiments done in a lab environment to test the proposed method. The results are analyzed and discussed in Section~\ref{sec:discussion}, and Section~\ref{sec:conclusion_and_future_work} concludes the paper.



\section{RELATED WORK}
\label{sec:related_work}

Although numerous studies are devoted to the detection and grasping of tomatoes\cite{single_tomato_1, single_tomato_edge, single_tomato_depth, single_tomato_grayscale_otsu, single_tomato_mask_cnn, single_tomato_harvesting_object_detection, 6d_tomato, tomato_color_segmentation_hand, tomato_color_segmentation_otsu,
cluster_object_detection, cluster_harvesting_rgb, 6d_tomato_truss, 6d_tomato_truss_2, tomato_learned_segmentation, de_haan, picknpack}, the majority focuses on the harvesting or grasping of single tomatoes instead of trusses. Grasping the entire truss by just one tomato results in a high chance of the stem detaching from the tomato, which makes all these methods infeasible. Instead, the truss must be grasped by the stem (called the peduncle).

The first step to finding suitable grasp poses commonly relies on identifying the stem, which is considered a segmentation problem. Common methods are based on color, where thresholds are set by hand \cite{tomato_color_segmentation_hand} or with the use of adaptive thresholding methods, like Otsu \cite{tomato_color_segmentation_otsu} or k-means thresholding \cite{de_haan}. Color-based methods usually achieve poor results in varying lighting conditions and cannot cope with cluttered environments.

More recently, deep learning methods have been proposed. Rong et al.~\cite{tomato_learned_segmentation} use a YOLO (You Only Look Once) network to first identify tomato trusses in an image, and then a second YOLO network produces masks for the part of the stem above the tomatoes where grasping and cutting is feasible. Although good results have been reported, the downside of this approach is that generating segmentation masks for training is time and labor intensive.

Other recent methods include deep learning-based 6D-pose estimation. Kim et al.~\cite{6d_tomato} find the 6D-pose of tomato and stem to harvest individual tomatoes by using a known 3D model. Such a model would likely not work for truss tomatoes since they have more variance in their appearance. Also, this method requires training within a simulator to determine ground-truth 6D poses.

Zhang et al.~\cite{6d_tomato_truss, 6d_tomato_truss_2} focus on trusses where the 6D-pose is found using keypoints. The tomato trusses are modeled with 11 keypoints: six for the tomatoes and five for the peduncle and stem.  However, having a fixed number of keypoints is unsuitable when dealing with different types of trusses, which can have varying numbers of tomatoes.
Furthermore, many of the methods used in harvesting are not directly applicable when it comes to grasping out of a crate. This is because they rely on the assumption that the trusses in a harvesting environment are hanging vertically. Then, the grasping position can be chosen on the stem above the highest tomato and at an angle perpendicular to the direction of gravity. This assumption does not hold for grasping trusses from a crate.

To grasp trusses in a horizontal position, de~Haan~et al.~\cite{de_haan} proposed a graph-based method that finds a grasp pose closest to the calculated center of mass along the peduncle with sufficient space from junctions; the locations where pedicels are attached to the peduncle. The grasp angle is chosen to be perpendicular to the peduncle.
After segmenting the stem, the peduncle is found under the assumption that it is the longest path on the graph with a limited curvature. This method is a claimed improvement over the method by Gray and Pekkeriet \cite{picknpack}, which uses a random sample consensus (RANSAC) regressor to identify the peduncle by assuming that it makes up the longest continuous area present in the stem segment. Although this method could suffice, this approach ignores other parts of the stem and tomatoes, such as the calyxes. This results in grasp failures, as reported by the authors. Also, this method of finding the peduncle is sensitive to its hyperparameters and fails for oddly shaped trusses. Lastly, this method requires segmentation of the stem. For good performance in clutter and varying lighting conditions, stem segmentation should be based on deep learning, which again requires extensive labeling to train. None of the above methods is capable of learning from the success or failure of the executed grasp.

\section{PERCEPTION}
\label{sec:perception}

We assume a setting where tomato trusses are stacked on top of each other with the peduncle facing upwards, which is how they are commonly stored in a crate after harvesting. The manipulator's end-effector with the Intel Realsense D405 RGB-D camera is initially positioned approximately $0.75\pm0.1$\,m above the tomatoes to fit the entire crate in the camera's field of view. The problem is a top-down grasping problem, where the grasp pose is defined in 4D: the 3D position and the wrist roll angle (in the sequel also called the orientation).

\begin{figure}[htbp]
    \centering
    \includegraphics[width=0.5\columnwidth]{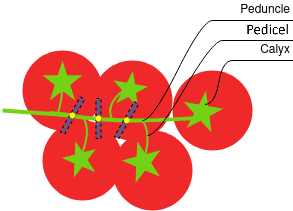}
    \caption{\small Suitable grasp poses on the peduncle for grasping a tomato truss. The yellow dots represent the positions, and the purple rectangles indicate the orientations of the grasps. }
    \label{fig:potential_grasp_poses_terminology}
\end{figure}

A major challenge in grasping tomato trusses is the requirement of a more accurate grasping pose than with many other simpler and less delicate objects. With parallel grippers, damage typically occurs when grasping by a tomato or the weaker parts of the stem, such as the pedicels or calyxes. A suitable grasp pose should be located on the peduncle, with as much space as possible from the other parts of the stem and tomatoes, as shown in Fig.~\ref{fig:potential_grasp_poses_terminology}. Also, when identifying a suitable grasping pose, it is crucial to consider obstructions. A truss should not be grasped if it is overlapped or obstructed by other trusses.\vspace*{-.5cm}

\begin{figure*}[htbp]
    \centering
    \vspace{0.2cm}
    \includegraphics[width=\textwidth]{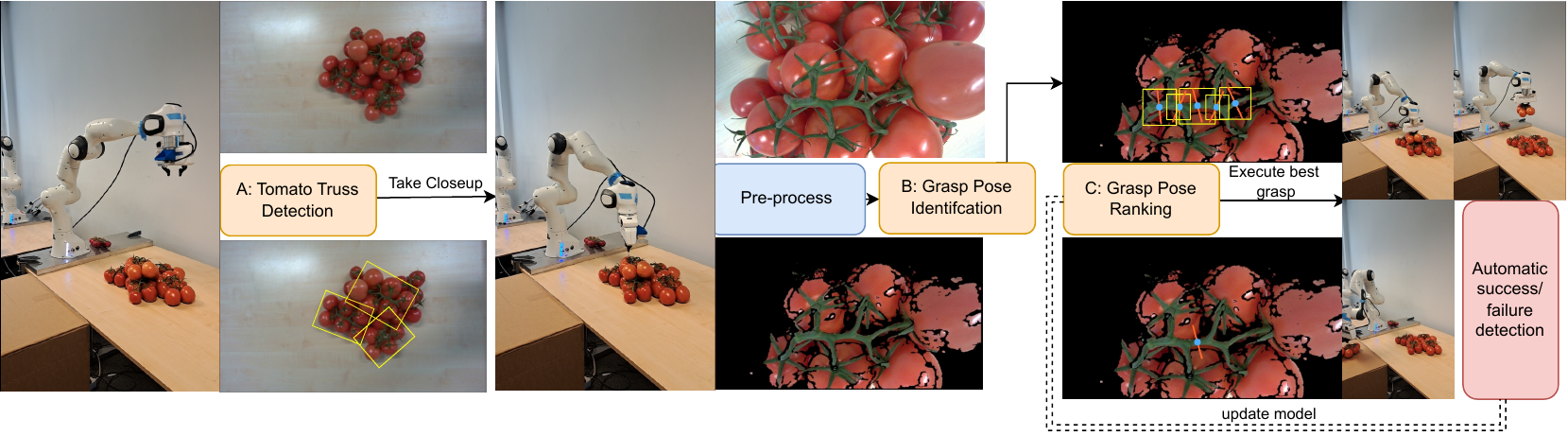}
    \caption{\small Overview of the method. First, the truss to be grasped is detected (steps A). The robot arm then moves the end-effctor with camera above this truss to take a close-up image, in which suitable grasp poses on the peduncle are identified (step B). Finally, the grasp pose ranking algorithm finds the most suitable pose (step C) and the robot executes the grasp. Based on the the grasp success or failure, the ranking model is adapted (dashed line in step C).}
    \label{fig:pipeline}
\end{figure*}

To find suitable grasp poses, we propose a three-stage perception method consisting of A) tomato truss detection, B) grasp pose identification, and C) grasp pose ranking. The individual stages are discussed below and visualized in Fig.~\ref{fig:pipeline}.

%
%
%

\subsection{Tomato Truss Detection}
\label{subsec:tomato_truss_detection}
This step aims at finding an unobstructed tomato truss. This is achieved by training a detection model on data in which only unobstructed trusses are labeled. When multiple such trusses are detected, the algorithm selects the one with the lowest average depth from the camera's perspective.

\paragraph{Architecture}
\label{tomato_truss_detection_archtecture}
We utilize a variant of the YOLOv5\cite{yolov5} architecture which outputs an oriented bounding box defined by the coordinates of all its corners $(x_1, y_1, x_2, y_2, x_3, y_3, x_4, y_4)$.
This format provides a precise fit, which is beneficial in the cluttered environment considered.

\paragraph{Dataset}
\label{paragraph:tomato_truss_detection_dataset}
We collected and labeled 225 images to train the model: 200 were used for training and 25 for validation. The images contained varying numbers and types of tomato trusses, and we varied the height and the angle from which the image was captured, as well as the background and lighting conditions. We have not recorded the exact variations, however, we have chosen them to approximately cover the robot's normal operating conditions. All images were resized to 640x640, and black borders were added if needed to preserve the aspect ratio.

\paragraph{Training}
\label{paragraph:tomato_truss_detection_training}
The network was trained for 300 epochs with a learning rate of 0.001 and a batch size of 32. The Adam optimizer was used with a momentum of 0.937 and weight decay of 0.0005. To reduce overfitting and improve generalization, the model weights were pre-trained on the COCO 2017 (Common Objects in\ Context) dataset\cite{coco}. We also applied common data augmentation techniques such as variations in the HSV (hue, saturation, value)\ channels, random rotation, translation, and scaling, and flipping upside-down and left-right.

\paragraph{Performance Evaluation}
\label{paragraph:tomato_truss_detection_evaluation}
The model performance evaluated on the validation set resulted in Mean Average Precisions MaP@0.5 of 0.952 and MaP@0.5:0.95:0.05 of 0.693.
During inference, the Non-Maximum Suppression (NMS) confidence and NMS IOU (Intersection over Union) thresholds were kept at the standard values of 0.25 and 0.45, respectively. This resulted in the precision of 0.935, the recall of 0.967, and the F1 score of 0.95.

\subsection{Grasp Pose Identification}
\label{subsec:grasp_pose_detection}
The next step is to identify candidate grasp poses on the tomato truss detected. We use a learning-based pose-estimation model, which takes an RGB image as its input and directly outputs the candidate grasp poses without the need for segmentation.

To get more precise depth information, we first control the robot arm to approach the truss to get a close-up view. 
The position of the arm is chosen so that the camera is 0.1\,m above the center of the bounding box. The orientation $\alpha$ of the camera is calculated so that it aligns horizontally with the longest side of the bounding box:
$$
    \alpha = \arctan2\left(y_2-y_1,x_2-x_1\right)
$$
where $\arctan2$ is the four-quadrant inverse tangent.

\paragraph{Preprocessing}
The closeup view mostly contains a single truss. However, this truss is usually still surrounded by parts of other trusses and parts of the underlying trusses can also be seen in the background. Therefore, two preprocessing steps are applied to the point cloud generated from the RGB-D image:
1) filter out the surrounding trusses by reusing the previously found bounding box of the truss of interest to remove points that lay outside the bounding box; 2) remove the background trusses by fitting a plane to the points remaining after step 1 by using the RANSAC method and removing points that have a distance larger than $d_p$. A suitable value for $d_p$ depends on the size of the tomatoes and is generally in the same range as their diameter. During experiments, $d_p=0.05$\,m was used.

\paragraph{Architecture}


To identify possible grasp poses on the preprocessed RGB image, we modified the Yolov7-Pose\cite{yolov7} architecture to detect for each potential grasping pose a bounding box containing a single keypoint and the respective orientation of the gripper. To get a 3D position for the grasp poses, the pixel locations of the keypoints are deprojected, and the grasp angles are taken directly as the keypoint orientations.

\paragraph{Dataset and Training}
A dataset of 50 preprocessed images was gathered and hand-annotated to train the model. This dataset was split into 40 training and 10 validation images. The same data augmentation and hyper-parameters were used as for the tomato truss detection model described in Section~\ref{subsec:tomato_truss_detection}.

\paragraph{Performance Evaluation}
The performance of the model is evaluated by comparing the keypoint predictions with the manually annotated ones (ground truth). A keypoint is considered correctly predicted if the distance to the ground truth is less than 0.003\,m, which is a little less than half the typical distance between junctions on the peduncle. Figure~\ref{fig:grasp_pose_detection_prediction} shows an example image with ground truth and predictions. A precision and recall score of 0.89 and 0.98, respectively, were obtained for the validation set. 
Figure~\ref{fig:keypoint_error} shows a box plot of the location and angle errors of correctly predicted keypoints.
\begin{figure}[htbp]
    \centering
    \includegraphics[width=0.9\columnwidth, height=2in,keepaspectratio]
    {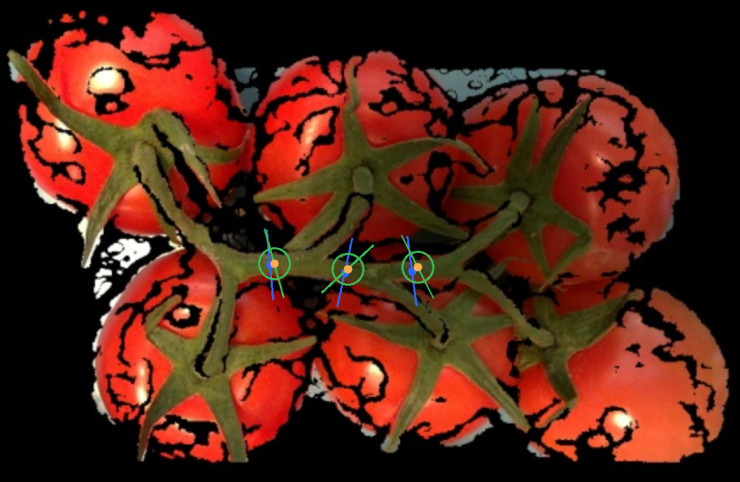}
    \caption{\small Example of the grasp pose identification network evaluation. The ground truth grasp poses are shown as orange dots with a green line for the orientation whilst the predictions are shown in blue. The green circles show the distance threshold in which the position prediction has to be located to be considered correct.}
    \label{fig:grasp_pose_detection_prediction}
\end{figure}
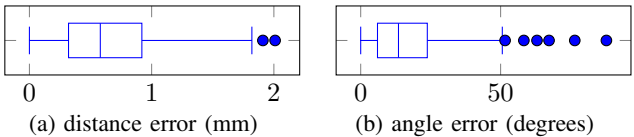
\begin{figure}[htbp]
    \centering
    \begin{subfigure}{0.49\columnwidth}
    \centering
        \begin{tikzpicture}
          \begin{axis}
            [
            ytick={1},
            yticklabels={},
            width=1.3\linewidth, 
            height=2.5cm, 
            enlarge y limits={0.5},
            ]
            \addplot+[
            boxplot prepared={
              median=0.58,
              upper quartile=0.32,
              lower quartile=0.92,
              upper whisker=0,
              lower whisker=1.82
            },
            ] coordinates {};
            \addplot+[
              only marks,
              mark=*,
              mark options={
                scale=1, 
                fill=blue, 
                draw=black
              }
              ]
              coordinates {
                (1.91,1)
                (2.01,1)
              };
          \end{axis}
        \end{tikzpicture}
    \captionsetup{skip=1pt} 
    \caption{distance error (mm)}
    \end{subfigure}
    \hfill
    \begin{subfigure}{0.49\columnwidth}
    \centering
        \begin{tikzpicture}
          \begin{axis}
            [
            ytick={1},
            yticklabels={},
            width=1.3\textwidth, 
            height=2.5cm, 
            enlarge y limits={0.5},
            ]
            \addplot+[
            boxplot prepared={
              median=13.5,
              upper quartile=6.0,
              lower quartile=23.8,
              upper whisker=0,
              lower whisker=50.6
            },
            ] coordinates {};
            \addplot+[
              only marks,
              mark=*,
              mark options={
                scale=1, 
                fill=blue, 
                draw=black
              }
              ]
              coordinates {
                (87.8,1)
                (63.0,1)
                (51.6,1)
                (67.3,1)
                (76.5,1)
                (58.3,1)
              };
          \end{axis}
        \end{tikzpicture}
    \captionsetup{skip=1pt} 
    \caption{angle error (degrees)}
    \end{subfigure}
\caption{\small Boxplots displaying the distance and angle errors of the correctly predicted keypoints on the validation set of the grasp pose identification network.}
\label{fig:keypoint_error}
\vspace{-0.2cm}
\end{figure}

\subsection{Grasp Pose Ranking}
\label{subsec:grasp_pose_ranking}
The last step in the perception method is the ranking of the identified grasp poses in terms of the expected grasp success. De~Haan et al.~\cite{de_haan} chose the grasp pose to be as close as possible to the estimated truss' center of mass to retain the truss's horizontal position after lifting it up. However, this method fails to account for possible collisions of the gripper with the pedicels or tomatoes.
To overcome this limitation, we propose a model that estimates the suitability of a grasping pose by a number between 0 and 1. To prevent the repetition of unsuccessful grasp attempts, this model is continuously updated online. In this way, the failure of recent grasp attempts leads to trying alternative grasp poses.

To get the input for the ranking method, for every possible grasping pose, the preprocessed point cloud gets rotated by the grasping angle, and points with an $L_{\infty}$ distance of more than $d_r$ to the grasping position get removed. This distance $d_r$, should be chosen so that all necessary local information remains. During experiments, this was set at $0.02$\,m. Finally, these resulting point clouds get projected back into the depth images, which are normalized and have a resolution of $128\times 128$ pixels.

\begin{figure*}[ht]
    \centering\includegraphics[width=.8\textwidth]{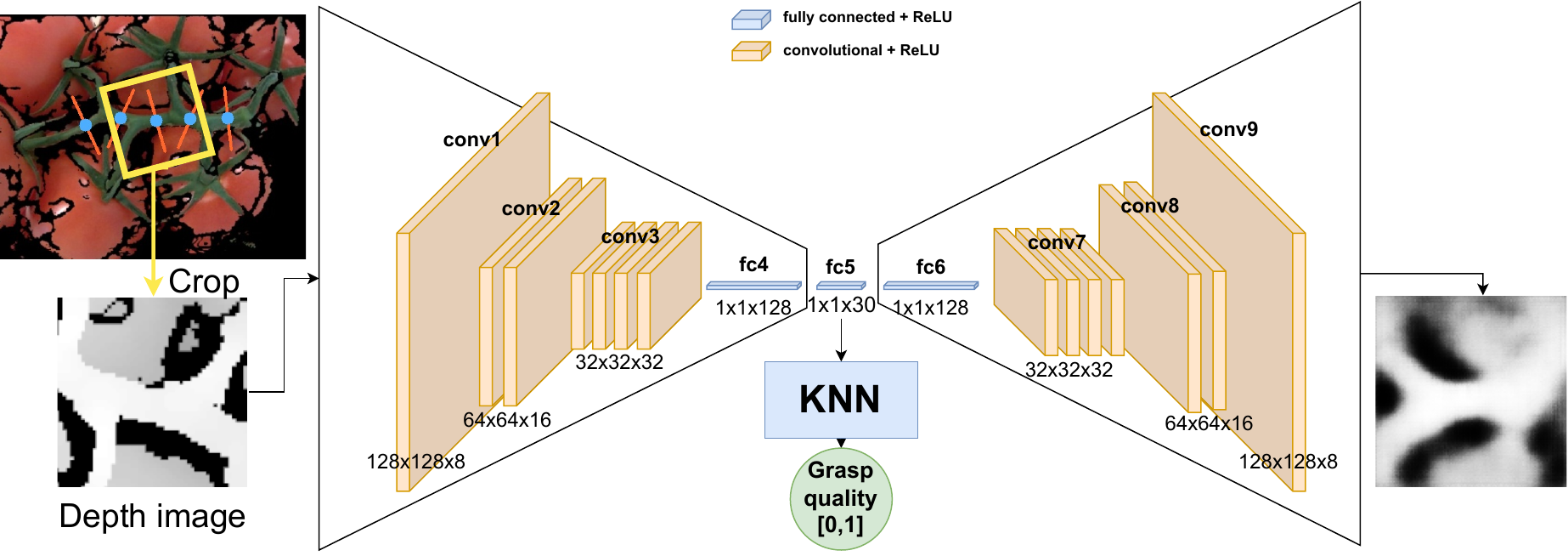}
    \caption{Architecture used for grasp pose ranking; a KNN classifier on the latent space of an auto-encoder.}
    \label{fig:autoencoder_knn}
\vspace*{-.5cm}
\end{figure*}

To be able to quickly adapt the model to new data, we use a KNN (k-nearest neighbors) classifier applied to features extracted by an auto-encoder, as depicted in Fig.~\ref{fig:autoencoder_knn}. While the auto-encoder is not updated online, the KNN classifier can easily be adapted in real time, even on a CPU. Utilizing an auto-encoder introduces an additional advantage over a direct neural network for the classification; although only one of the detected grasp poses can be executed and labeled, all the grasp proposals can be used for training the auto-encoder since this training does not require any labels.

\paragraph{Dataset}
To train the auto-encoder and KNN, an initial offline training phase is performed where one of the sampled grasping poses is randomly executed. Measuring the grasp success is automated by using the load force estimate provided by the manipulator; the change in force before and just after releasing the potentially grasped tomato truss is compared. The grasp is considered successful if this change is more than a predefined threshold. A total of 962 grasps on roughly 50 different trusses were recorded over multiple experiments within a period of several weeks. This resulted in 4807 grasp poses used to train the auto-encoder. This dataset was split into 70\% samples for training and 30\% samples for validation.

\paragraph{Training}
To train the autoencoder, we used the Adam optimizer with a learning rate of 0.0001, weight decay of 0.0001, and $\beta$ of 0.9. The model was trained for 40 epochs with a batch size of 512 using a mean squared error loss. For the KNN, the number of neighbors was set to 10, and the weights of each were set inversely proportional to their distance. For both steps, the dataset was augmented by using a combination of flipping upside-down, left-right and rotating by 180 degrees.

\paragraph{Performance Evaluation}
Of the 290 validation images, a total of 172 true positives (59.3\%), 72 true negatives (24.8\%), 14 false negatives (4.8\%), and 32 false positives (11.0\%) were obtained. This results in an F1 score of 0.88.
%

\section{EXPERIMENTS}
\label{sec:experiments}

To evaluate the proposed learning-based method, lab experiments were performed. An example video\footnote{\url{https://youtu.be/AnTaVYxz63c}} and the codebase\footnote{\url{https://github.com/LuukvandenBent/learning_approach_to_robotic_grasping_of_vine_tomatoes}} are available online.

\subsection{Experimental Setup}
\begin{figure}[htbp]
    \centering
    \begin{subfigure}[t]{0.24\columnwidth}
        \centering
        \includegraphics[width=\textwidth, height=1in, keepaspectratio]{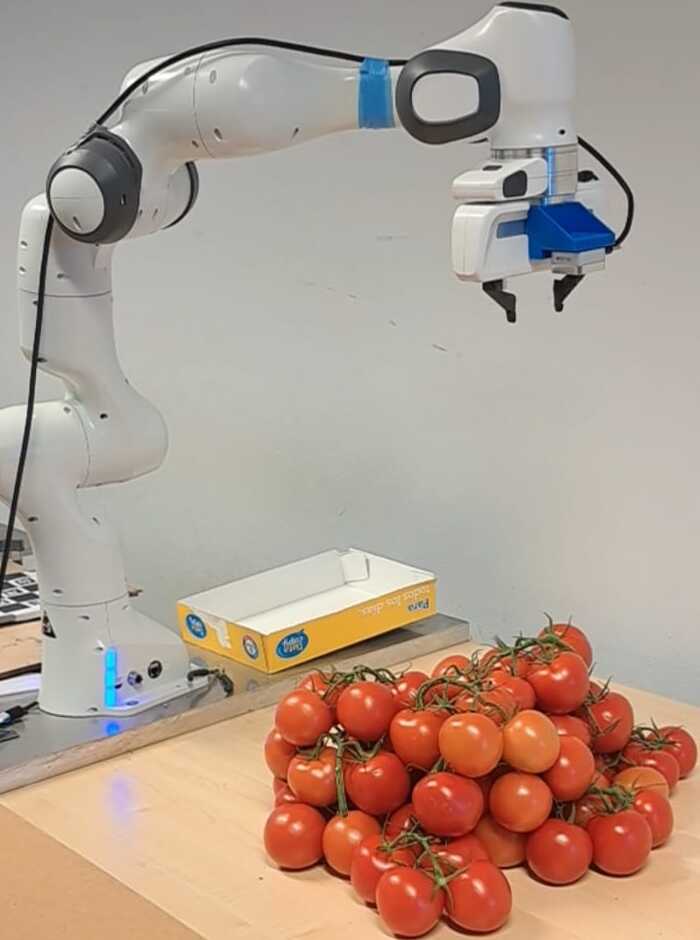}
        \caption{Setup}
        \label{fig:gripper_setup}
    \end{subfigure}
    \hfill
    \begin{subfigure}[t]{0.24\columnwidth}
        \centering
        \includegraphics[width=\textwidth, height=1in, keepaspectratio]{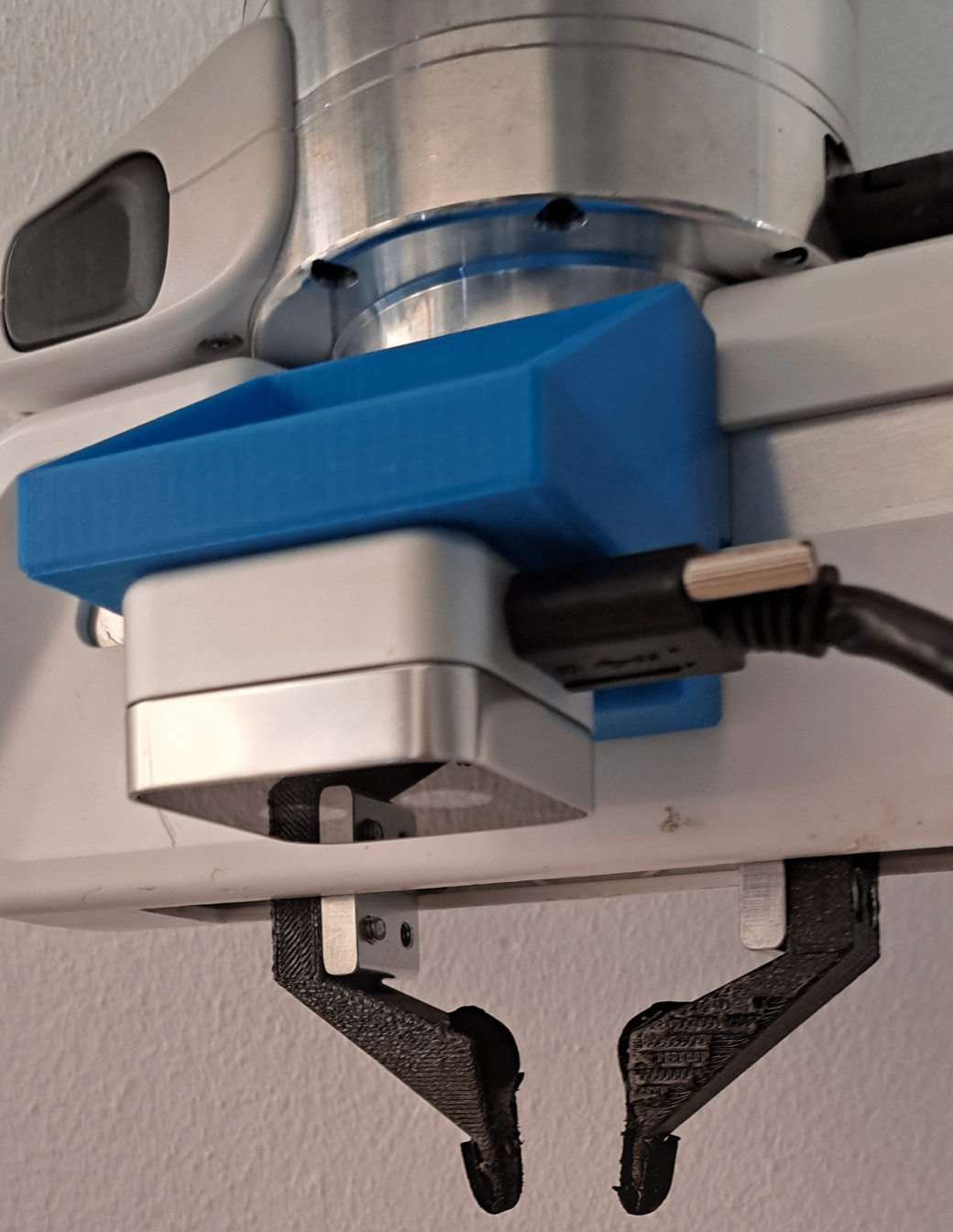}
        \caption{End-effector}
        \label{fig:gripper_on_robot}
    \end{subfigure}
    \centering
    \begin{subfigure}[t]{0.34\columnwidth}
        \centering
        \includegraphics[width=\textwidth, height=1in, keepaspectratio]{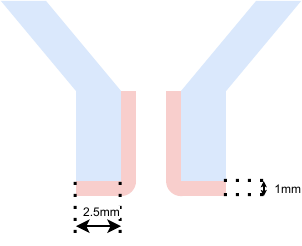}
        \caption{Front view}
        \label{fig:gripper_front}
    \end{subfigure}
    \hfill
    \begin{subfigure}[t]{0.14\columnwidth}
        \centering
        \includegraphics[width=\textwidth, height=1in, keepaspectratio]{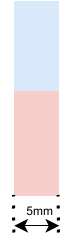}
        \caption{Side view}
        \label{fig:gripper_side}
    \end{subfigure}
    \caption{\small Experimental setup, with a closeup of the end-effector. The dimensions of the gripper fingers are shown from the side and front. The red area represents the deformable material.}
    \label{fig:setup}
\end{figure}

We used the Franka Emika Panda\footnote{\url{https://www.franka.de/research}} manipulator equipped with the Intel Realsense D405\footnote{\url{https://www.intelrealsense.com/depth-camera-d405/}} RGB-D camera, mounted close to the end-effector in the "eye-in-hand" configuration. Custom, 3D-printed slim gripper fingers accommodate the limited space available for grasping and lower the chance of the end-effector getting stuck on parts of the stem. The fingers were covered with grippy, deformable foam to minimize potential damage to the stem and tomatoes. A Cartesian impedance controller \cite{impedance_controller} is used to move the arm. The physical setup, along with a close-up of the end-effector and the dimensions of the fingers, can be seen in Fig.~\ref{fig:setup}.

The setup is mounted on a table where one or more tomato trusses are placed and potentially stacked next and on top of each other, depending on the experiment. The trusses were not placed inside a crate since we focus on perception and do not consider collision with the crate walls. We assume that i) the peduncle is facing upward (this is normally the case in practice), ii) the perception system performance is not influenced by the presence of the crate (this can be ensured by proper lighting using special lamps, which is a common industrial practice), and iii) the controller perfectly executes the commanded movement of the arm (which again is a reasonable assumption with current industrial robot arms).

\subsection{Pick and Place Routine}
To evaluate the method, a pick-and-place routine is carried out. It consists of five consecutive steps: (i) localizing a single truss to be grasped, (ii) approaching the truss with the eye-in-hand camera to take a close-up image, (iii) identifying a suitable grasping pose, (iv) reaching towards the peduncle and grasping it, (v) lifting the truss and placing it at a desired location.

We have conducted three types of experiments to evaluate how well the proposed method is able to grasp previously unseen tomato trusses:
\begin{enumerate}[]
  \item Grasping in a non-cluttered environment. Here, only one truss was placed on the table at a time. The goal of this experiment is to evaluate the impact of the proposed grasp pose ranking method, compared to randomly selecting one of the candidate poses or selecting the one closest to the center of the truss' bounding box. The center of the bounding box is an approximation of the center of mass, which should provide the most stable grasp.

  \item Grasping in a cluttered environment. Trusses are arranged to form a single underlying layer on top of which a single target truss placed. Here, the tomato truss detection and preprocessing part of the perception process are tested to see how well they are capable of dealing with the cluttered environment.
  \item Pile clearing. Lastly, multiple trusses are randomly stacked next to and on top of each other, which resembles a filled crate after harvesting. In this test, the purpose is to see if the system is able to one-by-one grasp all trusses that are present and not get stuck by repeatedly trying the same failing grasp pose.
\end{enumerate}

In the first two experiments, if a grasp is successful, the truss is placed back on the surface with a random pose near the center of the workspace by the manipulator. However, if a grasp fails, the truss is not moved by hand and is reattempted as is. The online learning capabilities of the proposed method are disabled to evaluate the performance of the model trained offline. Only in the pile-clearing experiment the online learning capability is enabled.

\subsection{Failure Modes}
Two types of failures were observed during experiments:
\begin{enumerate}[]
  \item Perception: the perception system provided an inappropriate grasping pose.

  \item Gripper: the truss slips out of the fingers during lifting or mid-air manipulation. This error indicates weakness in the pinch grasping method used.
\end{enumerate}
The type of failure was automatically determined during the grasping attempt by checking the width between the fingers after closing but before lifting. A width of (near) zero indicates that there is nothing between the fingers, signifying a failed grasping pose. If something was initially held between the fingers before lifting but not when placing, the error is assumed to be caused by slipping.

\subsection{Results}
The previously described pick and place tasks were executed on 25 different tomato trusses. For the first non-cluttered experiment, each truss was attempted with the three strategies: randomly picking a candidate grasp pose, picking the one closest to the center of the bounding box, and using the highest-scored pose ranked by the proposed method. Per truss, each strategy was repeated 20/10/10 times for the three methods, respectively, for a total of 1000 attempts. The outcomes displayed respective failure rates of 47.6\% (238/500), 20\% (50/250), and 7.2\% (18/250) and are summarized in Tab.~\ref{table:results_heuristic}.

In the remaining experiments, only the proposed grasp pose ranking strategy is used. In the cluttered-environment experiment, all 25 different tomato trusses were tested and repeated 10 times each. For each truss, a new single layer of the underlying trusses was formed using a random selection of the remaining trusses. A failure rate of 4.4\% (11/250) was observed, as shown in Tab.~\ref{table:results_clutter}.

The pile-clearing experiment was also performed 10 times. Each time, 10 out of the 25 tomato trusses were randomly selected and stacked by hand next to and on top of each other. In this experiment, the online learning of the grasp pose ranking method was enabled, but we removed the samples from the last attempts after each trial to make sure the trusses were unseen by the system at the beginning of each trial. Unlimited grasping attempts were allowed until the system was able to fully clear the pile in all 10 attempts. 93\% (93/100) of the trusses were successfully grasped at the first attempt, 6\% took two attempts, and one truss took six attempts.
%
%
\begin{table}[htbp]
\vspace{0.2cm}
\caption{\small Failure rates in the non-cluttered environment experiment (Experiment~1)}
\centering
\setlength{\tabcolsep}{4pt} 
\begin{tabularx}{\columnwidth}{lllll}
\hline
\begin{tabular}[l]{@{}l@{}}grasp pose\\selection \end{tabular} & \begin{tabular}[l]{@{}l@{}}trials\\ trusses*attempts\end{tabular} & total failures & gripper & perception \\
\hline
random & $25*20=500$ & 238 (47.6\%) & 117 (23.4\%) & 121 (24.2\%) \\
center & $25*10=250$ & 50 (20.0\%) & 35 (14.0\%) & 15 (6.0\%) \\
ranking & $25*10=250$ & 18 (7.2\%) & 13 (5.2\%) & 5 (2.0\%) \\
\hline
\end{tabularx}
\label{table:results_heuristic}
\end{table}
\begin{table}[htbp]
\caption{\small Failure rates of ranking-based method for grasping in isolation or clutter (Experiments 1 and 2, respectively)}
\centering
\begin{tabularx}{\columnwidth}{Xllll}
\hline
\multicolumn{1}{l}{scenario} & \multicolumn{1}{l}{\begin{tabular}[l]{@{}l@{}}trials\\ trusses*attempts\end{tabular}} & total failures & gripper & perception \\ \hline
isolated & $25*10=250$ & 18 (7.2\%) & 13 (5.2\%) & 5 (2.0\%) \\
clutter & $25*10=250$ & 11 (4.4\%) & 9 (3.6\%) & 2 (0.8\%) \\
\hline
\end{tabularx}
\label{table:results_clutter}
\end{table}

\section{DISCUSSION}
\label{sec:discussion}

The results of the first experiment in the non-cluttered setting show that the proposed ranking-based method achieves a lower failure rate compared to selecting the pose randomly or close to the truss center of mass.

In the second experiment, a lower failure rate was obtained when grasping tomato trusses in the cluttered setting compared to isolated trusses. Intuitively, the latter should be an easier task, which likely means that the difference is not statistically significant. However, these results show that the tomato truss detection and preprocessing steps effectively simplify the cluttered problem to essentially grasping in isolation. Most perception failures were a result of inadequate proposals by the grasp pose identification network. Therefore, the overall performance could be further improved by increasing the amount of data used to train this network.

After the experiments, no visual damage was observed on the tomatoes. However, after repeatably gripping and releasing the same trusses, some abrasion damage was observed on the skin of the peduncles. This will clearly not be an issue in the industrial setting, where each truss will be handled only once and, in addition, a specialised gripper can be used.

A limitation of our validation experiments is that they were performed with only one tomato variety. Although the system was during its development tested on other tomato varieties, no reliable conclusions can be drawn about the success rate of the proposed method for other varieties, for instance, cherry tomatoes.




The perception steps, tomato truss detection, preprocessing, grasp pose identification, and grasp pose ranking take approximately $0.05\pm0.01$, $2.00\pm0.24$, $0.64\pm0.21$, and $5.15\pm0.60$ seconds, respectively, on a computer with an Intel i7-8750H processor and NVIDIA GeForce GTX 1060 GPU. Note that the grasp pose ranking autoencoder was run on the CPU due to GPU memory constraints. The whole pick-and-place cycle, including returning the manipulator to the initial pose, takes around 30 seconds.

\section{CONCLUSION AND FUTURE WORK}
\label{sec:conclusion_and_future_work}

This paper presents a learning-based method for grasping tomato trusses in a cluttered environment. First, we use an object-detection model to identify an unobstructed truss to be grasped. Next, we employ an extension of the Yolov7-pose detection algorithm to allow the pose angle to be learned along with the keypoints, which is used to identify candidate grasping poses. In the last step, an autoencoder with a KNN classifier is trained offline and updated online to select the most promising grasp pose, based on the success of previous similar grasping attempts. Pile-clearing experiments conducted on a physical setup using real tomato trusses demonstrated a clearance rate of 100\% when allowed to retry after a failed attempt. Of all the trusses, 93\% were successfully grasped on the first try, while the remaining 7\% required more attempts.


Since many of the grasping failures were a result of the peduncle slipping out of the fingers, future work should focus on specialized grippers that can more effectively grip the trusses while avoiding damage to the stem. We further hypothesize that an enhanced gripper design can improve the grasp success also when attempting sub-optimal grasp poses, so reducing the influence of perception errors. Another topic for future research is the extension of our work to become collision-aware, solving the original problem of grasping out of a crate.

While the current models have been specifically trained for picking tomato trusses, the proposed method can be applied to a wide range of other objects. We verified this by testing it on bananas and silverware, which included knives, forks, and spoons. The preliminary results were promising and will serve as a basis for our future publications.

\noindent{\bf Acknowledgements}: This research was co-funded by the Netherlands Organization for Scientific Research project Cognitive Robots for Flexible Agro-Food
Technology, grant P17-01, and by the European Union under the project Robotics and Advanced Industrial Production (reg. no. CZ.02.01.01/00/22\_008/0004590).

\bibliographystyle{ieeetr}
\balance
\bibliography{references}

\end{document}